\documentclass[10pt,twocolumn,letterpaper]{article}

\usepackage{iccv}
\usepackage{times}
\usepackage{epsfig}
\usepackage{graphicx}
\usepackage{amsmath}
\usepackage{amssymb}
\usepackage{subfigure} 
\usepackage{color}

\usepackage[pagebackref=true,breaklinks=true,letterpaper=true,colorlinks,bookmarks=false]{hyperref}

\iccvfinalcopy 

\def\iccvPaperID{2097} 

\ificcvfinal\fi

\begin{document}

\title{What Properties are Desirable from an Electron Microscopy Segmentation Algorithm}

\author{Toufiq Parag\\
HHMI Janelia Research Campus\\
Ashburn, VA\\
{\tt\small toufiq.parag@gmail.com}
}

\maketitle
\thispagestyle{empty}


\begin{abstract}
The prospect of neural reconstruction from Electron Microscopy (EM) images has been elucidated by the automatic segmentation algorithms. Although segmentation algorithms eliminate the necessity of tracing the neurons by hand, significant manual effort is still essential for correcting the mistakes they make. A considerable amount of human labor is also required for annotating groundtruth volumes for training the classifiers of a segmentation framework. It is critically important to diminish the dependence on human interaction in the overall reconstruction system.  This study proposes a novel classifier training algorithm for EM segmentation aimed to reduce the amount of manual effort demanded by the groundtruth annotation and error refinement tasks. Instead of using an exhaustive pixel level groundtruth, an active learning algorithm is proposed for sparse labeling of pixel and boundaries of superpixels. Because over-segmentation errors are in general more tolerable and easier to correct than the under-segmentation errors, our algorithm is designed to prioritize minimization of false-merges over false-split mistakes. Our experiments on both 2D and 3D data  suggest that the proposed method yields segmentation outputs that are more amenable to neural reconstruction than those of existing methods.

\end{abstract}

\section{Introduction}

One important task for neural reconstruction from Electron Microscopy (EM) is to extract the anatomical structure of a neuron by accurately assigning regions of EM images to corresponding cells. Due to the size and number of EM images typically required for a useful dense reconstruction, it is impractical to manually perform such task. Recent studies on neural reconstructions or connectomics~\cite{takemura13}\cite{helmstaedter13a} apply automated segmentation algorithms for determining cell morphology. The result of such an automated segmentation algorithm is not free of errors, which is why a reconstruction approach must either manually correct the mistakes made by these algorithms~\cite{takemura13}, or conform them to a skeleton representation generated earlier by hand~\cite{helmstaedter13a}.

In addition, there have been many notable works addressing one or multiple processes constituting an overall segmentation algorithm. Existing algorithms such as\cite{jain10cvpr}\cite{ciresan12}\cite{jurrus10mia}  for pixel classification;~\cite{liu12watershed}\cite{liu14} for effective generation of over-segmentation; \cite{parag14}\cite{jain11}\cite{andres12} for isotropic 3D supervoxel clustering;  ~\cite{vazquez11}\cite{funke12} for co-segmentation for anisotropic data report impressive performances on different kinds of EM datasets. Many of these novel approaches are motivated by the methods in natural image segmentation and evaluate output accuracy using error measures popular in computer vision literature, e.g., Rand Error (RE) of~\cite{jain11}, Variance of Information (VI) of~\cite{andres12}\cite{parag14}. 

Ideally, an automated segmentation should attain $100\%$ accuracy -- its output should be free of both types of segmentation errors, namely false merge (under-segmentation) and false split (over-segmentation).  However, it is not realistic to expect (near) $100\%$ accuracy in practice; given the performances of the existing state of the art algorithms, one can generally assume that their outputs need to be corrected afterwards. Then, from a connectomics point of view, a segmentation algorithm should be designed to minimize manual labor (or algorithmic complexity) required for correcting its output\cite{jain10opinion}. 

To the best of our knowledge, there has not yet been a study analyzing the effect of segmentation errors on the effort necessary to correct them. Although error quantities, such as Rand Error (RE)~\cite{jain11}, provide a coarse assessment of the mistakes an algorithm makes, they are unable to conclusively forecast the amount of work required for refinement. As an example, inaccurately combining two regions of sizes A and B would incur the same RE value as incorrectly splitting one region of size A+B into two parts. However, rectifying these two mistakes demands significantly different amount of work~\cite{chklovskii10}. The high RE of a false split of two large bodies, e.g., $A = B = 10000$, on a $512 \times 512$ image disproportionately penalizes the effort to correct such error.   

\begin{figure}
\vspace{-0.4cm}
\begin{center}
\subfigure[\scriptsize Plane 50]{\includegraphics[width=0.3\columnwidth, height=0.3\columnwidth]{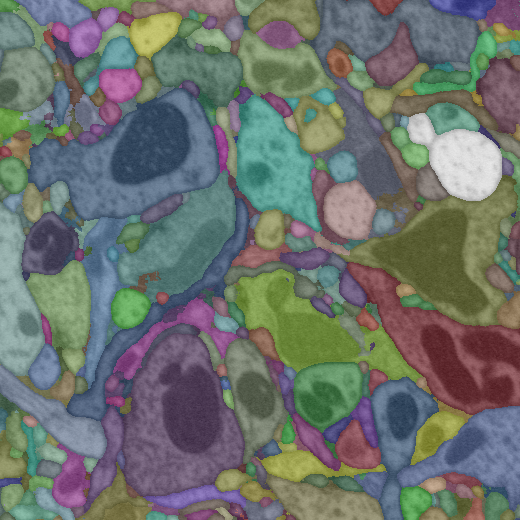}}
~\subfigure[\scriptsize Plane 220]{\includegraphics[width=0.3\columnwidth, height=0.3\columnwidth]{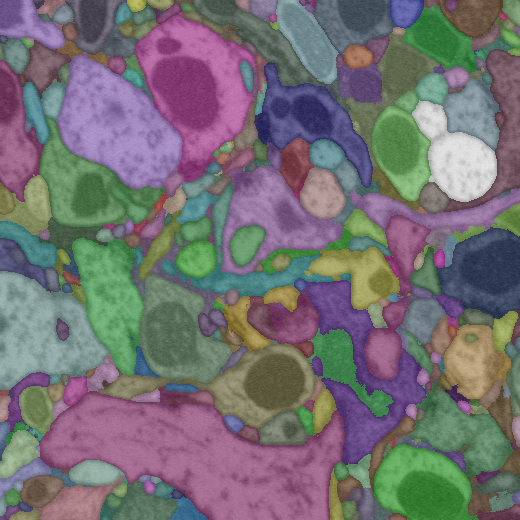}}
~\subfigure[\scriptsize Plane 484]{\includegraphics[width=0.3\columnwidth, height=0.3\columnwidth]{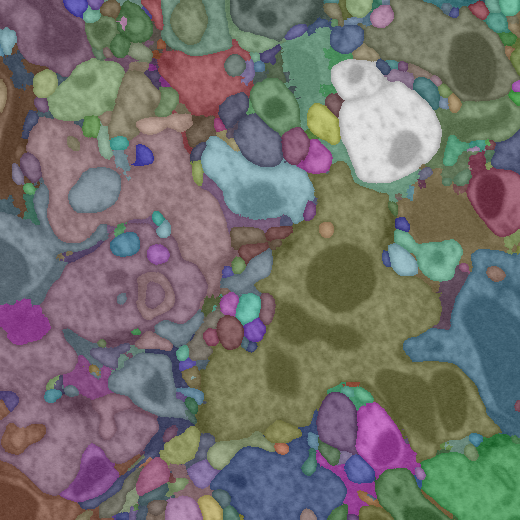}}
\vspace{-0.1cm}
\caption{\scriptsize 3D Segmentation output on 3 planes from a volume of 500 images. Each individual neuron has been colorized with a different color.  The two adjacent regions, colored in white, in the top-right, are in fact parts of two different neurons which have been falsely merged.  Manually correcting this under-segmentation error is much more labor-intensive than correcting a false split.}\label{F:INTRO}
\vspace{-0.7cm}
\end{center}
\end{figure}

From a reconstruction perspective, an over-segmented result is preferred over an under-segmented one because a fragmented set of regions can be refined by automated methods such as agglomeration~\cite{iglesias13}\cite{parag15b}\cite{parag14} or co-segmentation~\cite{funke12}, but an under-segmented region can only be fixed by a human expert. Even for a human expert, identifying and correcting false merges is more difficult than correcting false split~\cite{chklovskii10}. This difficulty is more pronounced in 3D volume segmentation than it is in 2D segmentation. Consider separating the two regions falsely connected through 450 planes (from 50 to 500) of a $520^3$ volume by a segmentation method as displayed in Figure~\ref{F:INTRO}. The authors of~\cite{iglesias13}\cite{parag15b}\cite{parag14} were aware of this issue and reported the two types of error rates separately for performance assessment. The study of ~\cite{jain10cvpr} attempts to reduce false merges by identifying the locations vital for preserving topology given exhaustive groundtruth of the data.  

Another desirable property of the EM segmentation algorithms is to be able to train the necessary components efficiently without compromising accuracy. An efficient training is perhaps essential for large scale reconstruction where one may anticipate learning the predictors multiple times for different neuropils. A quick segmentation result may also assist the neurobiologist to decide the optimal sample preparation that would maximize segmentation accuracy. But, training existing segmentation algorithms~\cite{jain10cvpr}\cite{ciresan12}\cite{jurrus10mia} remains a significant bottleneck in connectomics~\cite{helmstaedter13b} due to the time and effort necessary for generating the groundtruth and time complexity of training the classifier (e.g., deep neural networks). 
  
A highly curated exhaustive groundtruth, such as those offered by the segmentation challenges (e.g., ISBI 2012 2D, SNEMI 2013 3D), demands extensive effort. Provided necessary resources, it is possible to generate a reasonable groundtruth by iteratively refining segmentation on a small volume with an interactive labeling tool such as ilastik~\cite{ilastik11}. This label set is expected to contain a small degree of tolerable noise but is efficient to generate. Some recent algorithms~\cite{andres12}\cite{iglesias13}\cite{parag15b}\cite{parag14} have utilized interactively generated groundtruth to train the necessary tools for segmentation. However, these algorithms inherently rely on highly expert annotators or neurobiologists in order to produce a useful annotation efficiently (by finding out the minimal area to label for the prediction-correction scheme). Automated algorithms are expected to diminish such dependency on human expertise. As an alternative to exhaustive labeling, Jones et.al.~\cite{jones13sparse} presented a method for sparsely labeling the membrane locations based on appearance similarity to user annotated examples. A completely semisupervised approach like~\cite{jones13sparse} will be sensitive to the penalty parameter and has a risk of introducing noises that are too difficult for a classifier to tolerate.
\begin{figure}
\vspace{-0.4cm}
\begin{center}
\subfigure[\scriptsize]{\includegraphics[width=0.24\columnwidth, height=0.25\columnwidth]{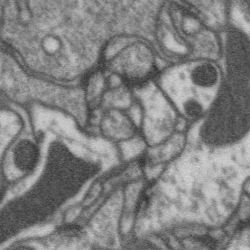}}
\subfigure[\scriptsize]{\includegraphics[width=0.24\columnwidth, height=0.25\columnwidth]{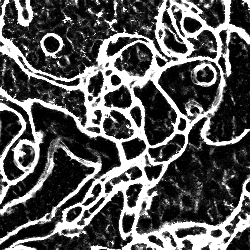}}
\subfigure[\scriptsize]{\includegraphics[width=0.24\columnwidth, height=0.25\columnwidth]{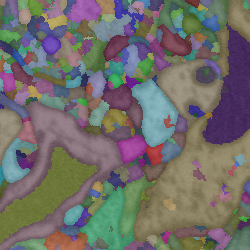}}
\subfigure[\scriptsize]{\includegraphics[width=0.24\columnwidth, height=0.25\columnwidth]{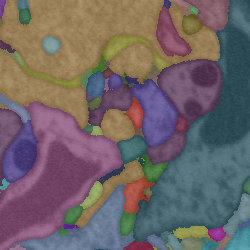}}
\vspace{-0.1cm}
\caption{\scriptsize Workflow of a standard EM segmentation framework: (a) input $\rightarrow$ (b) pixelwise classification (white: membrane, black: non-membrane) $\rightarrow$ (c) over-segmentation $\rightarrow$ (d) final segmentation.}
\label{F:PIPELINE}
\vspace{-0.7cm}
\end{center}
\end{figure}

We adopt a standard EM segmentation approach~\cite{jain11}\cite{andres12}\cite{iglesias13}\cite{parag14},  as illustrated in Figure~\ref{F:PIPELINE}, where the confidence values of a pixelwise classifier \footnote{\scriptsize We adopt multi-class pixel classification, as is explained later.} are utilized to generate an initial over-segmentation of the dataset. The over-segmentated image or volume is then refined by aggregating superpixels with the help of a superpixel boundary classifier. In this paper, we propose an algorithm for training pixel and superpixel boundary classifiers.  The classifiers are trained to attain two desirable properties of an EM segmentation method:

\noindent 1. Maximize efficiency: the proposed algorithm employs active learning for classification. Instead of requiring an exhaustive pixel-level groundtruth, \emph{our algorithm automatically determines a small fraction of samples that are critical for training} the pixel and superpixel boundary classifiers ($<1\%$ for pixel and $<20\%$ for superpixel boundary).  These examples are identified using the disagreement between two predictors: a) a classifier being updated iteratively, and b) a semisupervised label propagation algorithm~\cite{semi-super06} predicting labels based on feature similarity.  Unlike~\cite{jones13sparse}, all our training examples are labeled by an annotator.

\noindent 2. Minimize false-merge: without exhaustive groundtruth, it is not possible to locate the topologically critical pixels using the method of~\cite{jain10cvpr}. We hypothesize that emphasizing on the detection of membrane pixels over other types would reduce the amount of false merges. Accordingly, our training protocol is designed to be biased towards more accurate learning of membrane class than the remaining categories.

We empirically demonstrate the advantages of the proposed method over the state of the art techniques for neural reconstruction from both 2D and 3D EM data. The overall active learning algorithm is defined in Section~\ref{S:OVERALL}. Sections~\ref{S:ACTIVE_PIXEL} and~\ref{S:ACTIVE_SP}  explain how our active training approach is adapted for pixel and superpixel boundary classification. The following section (Section~\ref{S:EXP_RESULT}) discusses the experimental setup and reports the results. Finally, Section~\ref{S:DISCUSSION} concludes with a discussion summarizing our findings.

\section{Proposed Active Labeling Framework}\label{S:OVERALL}
The segmentation scheme we adopt consists of pixel classification followed by a superpixel clustering by means of a superpixel boundary classifier. We propose an active strategy to train both the pixel and superpixel boundary classifiers. The goal of an active learning method is to identify a few examples -- crucial for training a classifier -- from a pool of unlabeled samples. The proposed active classification scheme identifies the challenging examples from the dataset and requests their labels from user. Given the labels for the query examples, the algorithm reconfigures its predictors and identifies a new set of queries in a repetitive fashion. 

With the aim of locating these challenging examples, we estimate the class label of any unlabeled point by two predictors having substantially different views of the dataset. One predictor is a classifier (Random Forest (RF)~\cite{breiman01} in our experiments) trained from an initially available subset of datapoints $X_l \subset X = \{ x_1, \dots, x_n \}$  and their labels $Y_l$. The other predictor is a novel variant of semisupervised label propagation algorithm~\cite{zhu03}\cite{semi-super06}, that assumes a cluster formation of similar datapoints in feature space.  While the classifier assesses the class of an unlabeled example by a discriminative set of rules learned so far, the label propagation technique extrapolates a prediction based on feature similarity among the datapoints. 

\begin{figure}
\vspace{-0.4cm}
\begin{center}
\subfigure[\scriptsize Actual class labels]{\includegraphics[width=0.31\columnwidth, height=0.29\columnwidth]{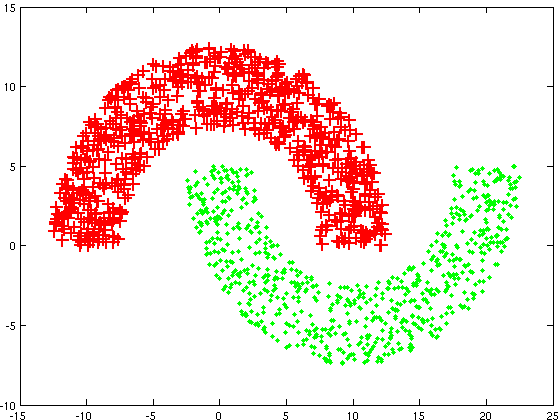}\label{F:MTHOD_EXPLN_INPUT}}
~\subfigure[\scriptsize Label Prop output]{\includegraphics[width=0.32\columnwidth, height=0.29\columnwidth]{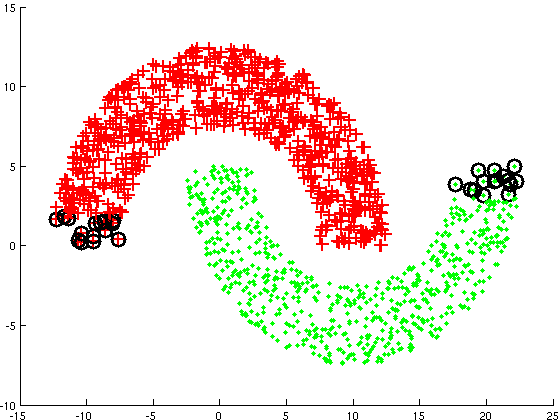}\label{F:MTHOD_EXPLN_LPROP}}
~\subfigure[\scriptsize RF prediction]{\includegraphics[width=0.32\columnwidth, height=0.29\columnwidth]{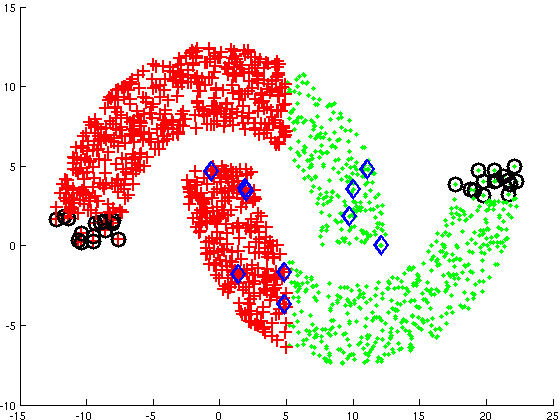}\label{F:MTHOD_EXPLN_RF}}
\vspace{-0.1cm}
\caption{\scriptsize \subref{F:MTHOD_EXPLN_INPUT} synthetic two-moon dataset with + and $.$ representing two classes; \subref{F:MTHOD_EXPLN_LPROP}, \subref{F:MTHOD_EXPLN_RF} predictions from label propagation and random forest classifier respectively given labeled examples circled in black. Blue diamonds mark new queries.}\label{F:MTHOD_EXPLN}
\vspace{-0.8cm}
\end{center}
\end{figure}

A training sample is considered to be challenging if the class suggested by feature similarity is different from that calculated by the discriminative rules and vice versa. For the interested readers, we illustrate the intuition behind our query generation approach on the synthetic two moon dataset on Figure~\ref{F:MTHOD_EXPLN}. Provided the same set of labeled examples, circled in black in Figure~\ref{F:MTHOD_EXPLN}\subref{F:MTHOD_EXPLN_LPROP} and~Figure~\ref{F:MTHOD_EXPLN}\subref{F:MTHOD_EXPLN_RF}, the label propagation can correctly extrapolate the labels of the rest of the datapoints utilized feature similarity (here euclidean distance between points) whereas a classifier, such as RF, will be unable to infer the class separation. Our method would select some samples, where the two predictions differ (marked by blue diamonds), as the next set of queries.

The disagreement among these two types of estimates is quantified by a ranking formula. The first few examples in descending order of disagreement measure are presented to the user as queries. The set $X_l$ is augmented by this new annotated queries and the whole process is repeated until a predefined stopping criterion is satisfied. 

In Section~\ref{S:MULTICLASS_LABEL_PROP}, we propose the semisupervised label propagation method for a multiclass setting to facilitate the multiclass approaches of~\cite{parag15b}\cite{iglesias13}. The strategies for query generation and initialization are different for pixel and superpixel boundary classification and are explained in Sections~\ref{S:ACTIVE_PIXEL} and~\ref{S:ACTIVE_SP}  respectively.  

\subsection{Proposed Multiclass Label Propagation}\label{S:MULTICLASS_LABEL_PROP}
Let us suppose, we have $n$ datapoints $x_i$ that we wish to classify into one of the $k$ classes. Let $\mathbf{f}_i$ denote the indicator variable for datapoint $x_i$: $f_i^c = 1$ if $x_i$ is classified to class $c$ and rest of its values are $0$. We wish to assign \lq similar\rq ~datapoints     into the same class, i.e., the pairs of samples $x_i$ and $x_j$ with large feature similarity quantified by $w_{ij}$ should belong to the same class. We propose to attain this by minimizing the following cost.
\vspace{-0.2cm}
{\small
\begin{eqnarray}  
J(\mathbf{f}) &=& \sum_{i \sim j} w_{ij}~ \biggl [ {\mathbf{f}_i \over \sqrt{d_i}} -  {\mathbf{f}_j \over \sqrt{d_j}} \biggr ]^T \biggl [ {\mathbf{f}_i \over \sqrt{d_i}} -  {\mathbf{f}_j \over \sqrt{d_j}} \biggr ] \\
&=& 2 \sum_i \mathbf{f}_i^T \mathbf{f}_i  - 2 \sum_{i \sim j } {w_{ij} \over \sqrt{d_i} \sqrt{d_j}} \mathbf{f}_i^T \mathbf{f}_j.
\vspace{-0.1cm}
\end{eqnarray}}

In this cost function, we normalize the weight $w_{ij}$ by the corresponding degree $d_i = \sum_j w_{ij}$ to balance the effects of disparity in class sample size. The cost is summed over all neighboring $i \sim j$ that possess a feature similarity above a certain predefined value.  Using a matrix notation for the indicator variables, $F = [ \mathbf{f}_1^T, \dots \mathbf{f}_n^T]^T$, we can write this cost function as 
\vspace{-0.2cm}
{\small
\begin{equation}
J(F) = 2 ~\text{Tr} \{ F F^T (I - D^{-0.5} W D^{-0.5})\},
\vspace{-0.1cm}
\end{equation}}
where $I$ and $D$ are the identity and diagonal degree matrices respectively.  By relaxing the values of $F$ to be nonnegative real-valued numbers $f_i^c \ge 0$ and differentiating wrt $F$, one can compute the system of linear equations needed to be solved for determining $F$. Of course, the minimization is constrained by label consistency among the values of $\mathbf{f}_i$, i.e., $F \mathbf{1} = \mathbf{1}$, where $\mathbf{1}$ is a vector of all 1's.
\vspace{-0.2cm}
{\small 
\begin{equation}\label{E:LABELPROPOBJ}
{\partial J \over \partial F} = 0 \implies \Bigl (I - D^{-0.5} W D^{-0.5} \Bigr) F = 0
\vspace{-0.1cm}
\end{equation}}

An efficient solver for Equation~\ref{E:LABELPROPOBJ} is essential to build an interactive interface of our method. By avoiding the factorization of matrices with thousands of variables, iterative techniques can produce a solution significantly faster than the closed form methods with the same level of accuracy ( up to a certain error tolerance). A stationary iterative formulation of this equation would repeatedly update the solution using the following formula~\cite{kelley95}.
\vspace{-0.2cm}
{\small
\begin{equation}\label{E:LABELPROPITR}
F_{next} = D^{-0.5} W D^{-0.5} F
\vspace{-0.1cm}
\end{equation}}
This iteration will converge if:  1) the absolute value of the eigenvalues of  $D^{-0.5} W D^{-0.5}$ is bounded by 1, and 2) $I-D^{-0.5} W D^{-0.5}$ is non-singular~\cite{kelley95}. Since there is no bipartite connected component in the graph corresponding to $W$, the first condition is satisfied~\cite{chung96}. We add a small perturbation to the quantity $D^{-0.5} W D^{-0.5}$ to attain non-singularity. One must also satisfy the label consistency constraint $F \mathbf{1} = \mathbf{1}$ to reach a meaningful solution. 

In our active learning setting, the algorithm is given the labels for $m$ out of $n$ examples (where $m << n$) at the beginning of the process. We set the known labels in $F$ and iterate Equation~\ref{E:LABELPROPITR} followed by a projection onto $F \mathbf{1} = \mathbf{1}$ until convergence for computing the unknown label confidences. The algorithm is outlined in Table~\ref{T:ALGORITHM} and has similarity to a past approach for efficient label propagation on large dataset~\cite{zhu02tech}.
\begin{table}
\small
\vspace{-0.3cm}
\caption{Multiclass Label Propagation algorithm}\label{T:ALGORITHM}
\begin{center}
\begin{tabular}{l}
\hline
Algorithm: Multiclass Label Propagation\\
  repeat \\
  \quad 1. Set the known labels in $F$.\\
  \quad 2. Update solution by Equation~\ref{E:LABELPROPITR}. \\
  \quad 3. Project onto $F \mathbf{1} = \mathbf{1}$ \\
  until convergence\\
\hline
\end{tabular} 
\end{center}
\vspace{-0.8cm}
\end{table}

After a query set is annotated by user, the linear equations in~\ref{E:LABELPROPOBJ} need to be solved again. Instead of starting the solver algorithm (Table~\ref{T:ALGORITHM}) from scratch, we begin with the most recently converged $F$ as the initial solution. Such a warm start brought about a significant speed-up without altering the output in our experiments. 

\subsection{Active Learning for Pixel Classification}\label{S:ACTIVE_PIXEL}
In pixel classification, each datapoint $x_i$ of the above formulation corresponds to a pixel. We will denote a pixel by a different literal $u_i$ to distinguish it from it from superpixel boundary defined later. In our design, each pixel is classified into one of the four classes: membrane, cytoplasm, mitochondria, mitochondria border~\cite{parag15b}. 

\noindent \textbf{Initial Subset Selection :} Equal size subsets of samples, one for each class, are selected from the dataset to constitute the initial dataset $X_l$ for label propagation. In the interactive setting, the user will be required to select the initial $X_l$ using a GUI.

In an attempt to maximize the detection of membrane pixels, the initial training set for the RF classifier is constructed from a subset of $X_l$ that contains different number of examples for different classes. In the following text, we describe how the pairwise similarity values in $W$ are utilized to determine the sample proportion for different categories. 

Introducing indicator vectors $\boldsymbol{\alpha}^m$ and $\boldsymbol{\alpha}^o$ for membrane class $m$ and other classes $o$ respectively, one can determine the sample proportion by solving an optimization problem.  The value of $\alpha^m_i = 1$ if the i-th membrane example in $X_l$ is selected and $\alpha^m_i=0$ otherwise. The following formulation will select of largest subset of initial samples that will prevent misclassification of any member of class $m$ in a nearest neighbor classifier setting. 
\vspace{-0.2cm}
{\small
\begin{align}\label{E:INITIAL_OPT}
\max_{\boldsymbol{\alpha^o}, \boldsymbol{\alpha^m}} ~&~\sum_i \alpha^o_i + \alpha^m_i \nonumber \\
\text{s.t.}~~ \sum_{l : y_l = m} \alpha^m_l w_{ij} \alpha^m_j ~&\ge~ \sum_{i : y_i = o} \alpha^o_i w_{ij} \alpha^m_j, ~~~ \forall_j~ y_j = m \nonumber  \\
\alpha^m_i, \alpha^o_i ~& \in \{0, 1\}
\vspace{-0.2cm}
\end{align}}

Here, $y_i \in \{m,o\}$ indicates whether $u_i$ belongs to membrane or other categories. In practice, we compute a sub-optimal solution to this problem for efficiency. In our solution, $\alpha^m_i = 1$ for all $i$.  We then greedily select examples for each class $o$ to increase~$\text{Cut}(m,o)$ as long as $\text{Vol}(o) \le \text{Vol}(m)$; we refer the reader to~\cite{meila01} for the definitions of these terms and to comprehend the motivation behind our heuristics. 

\noindent \textbf{Query Generation :} Let the vector $\mathbf{p}_i$ denote the prediction confidences generated by the classifier for an unlabeled pixel $u_i$, where $p_i^c$ corresponds to the confidence towards class $c$. If one wishes to compute the over-segmentation from the classifier probability for membrane class $p_i^m$, it is favorable to have $p_i^m > p_i^o,~ o \ne m$ for all membrane pixels. For a pixel $u_i$ from the other classes, the deviance $p_i^o - p_i^m$  should be maximized instead. We define a margin vector $\bar{\mathbf{p}}_i$ wrt class $m$ consisting of these quantities defined as follows. 
\vspace{-0.2cm}
{\small
\begin{align}\label{E:PIXEL_DEV}
a &= \arg\max_c p_i^c \nonumber\\
\bar{p}_i^c &=  \begin{cases}
        0,    & c \ne a \\
        p_i^m, & c = a = m \\
        p_i^a - p_i^m, & c = a \ne m
\end{cases} 
\vspace{-0.2cm}
\end{align}}

Let $\bar{\mathbf{g}}_i$ be the margin wrt class $m$ computed for $u_i$ in a similar fashion from the real-valued outputs of multiclass label propagation algorithm. The disagreement $\delta(u_i)$ between these two estimates is computed by the dot product of their differences.
{\small
\vspace{-0.2cm}
\begin{equation}
\delta_{\text{pixel}}(u_i) = (\bar{\mathbf{g}}_i - \bar{\mathbf{p}}_i)^T (\bar{\mathbf{g}}_i - \bar{\mathbf{p}}_i).  
\vspace{-0.1cm}
\end{equation}}
The margins $\bar{\mathbf{g}}_i$ and $\bar{\mathbf{p}}_i$ are modeled to capture the overlap in confidence the two predictors have between membrane and other classes. The disagreement $\delta_{\text{pixel}}(u_i)$ between these two margins will increase when the confidence distributions deviate from one another. A few unlabeled samples with largest disagreement value $\delta_{\text{pixel}}(u_i)$ will be selected as the next set of queries to be presented to the user. After the termination of the training process, the real-valued confidences of the classifier (RF in our case) are used for the subsequent tasks.

\subsection{Active learning for Superpixel Boundary Classification}\label{S:ACTIVE_SP}
The output confidence of pixel classifier (RF in our cases) is utilized to generate an over-segmentation of the image or volume (see Figure~\ref{F:PIPELINE}). In order to aggregate the fragments into actual cell regions, each boundary between two superpixels of this over-segmentation needs to be classified as true or false boundary. We employ a  superpixel boundary classifier (RF) that is also trained using the active learning method. For this training, each datapoint $v_i$  corresponds to a superpixel boundary. 

\noindent \textbf{Initial Subset Selection :} In order to reduce redundancy, the initial labeled set $X_l$ was populated by the centers of the output of a clustering algorithm such as k-means. 

\noindent \textbf{Query Generation :} Given the real valued confidences $q_i$ from the current classifier and the estimates $h_i$ of the label propagation method, we use the following formula to compute disagreement between them.
\vspace{-0.2cm}
{\small
\begin{equation}
\delta_{\text{sp}}(v_i) = (q_i - h_i)^2.
\vspace{-0.1cm}
\end{equation}}
Note that, since there are only two classes, values of both $q_i$ and $h_i$ are scalar for superpixel border classification. A few samples with largest $\delta_{\text{sp}}(v_i)$ are selected as the next query set to be annotated. After the training terminates, the real valued predictions from RF are used for superpixel clustering.

\section{Experiments and Results}\label{S:EXP_RESULT}

The proposed algorithm has been tested for both 3D volume and 2D image segmentation problems. In the following, we will describe the experimental setup, i.e., computation of the intermediate quantities, feature representation etc. for pixel and superpixel boundary classification. The Sections~\ref{S:RESULT3D} and~\ref{S:RESULT2D} report the results on 3D and 2D data respectively.  

\subsection{Experimental Setup}\label{S:EXPSETUP}
\noindent \textbf{Pixel classification :} As noted earlier, each pixel was classified into four classes: membrane, cytoplasm, mitochondria, and mitochondria border. A pixel is represented by features similar to those utilized in ilastik~\cite{ilastik11}, e.g., gaussian smoothing, gradient magnitude, laplacian of gaussian, hessian of gaussian and its eigenvalues, structure tensor and its eigenvalues etc. computed at different scales. The similarity values for a pair of examples $\{u_i, u_j\}$ were generated by gaussian distance between their feature representations:  $w_{ij} = \exp\bigl \{ -{1 \over 2} (\boldsymbol{\phi}_i - \boldsymbol{\phi}_j)^T \Sigma^{-1} (\boldsymbol{\phi}_i - \boldsymbol{\phi}_j) \bigr \}$   where $\boldsymbol{\phi}_i$ are the feature values of $u_i$ and $\Sigma$ is the covariance matrix among all feature vectors.

\noindent \textbf{Superpixel boundary classification :}
Given the pixel detection result, we utilize the predicted confidence values of the membrane class for generating an over-segmentation by the watershed algorithm~\cite{meyer93}. In order to generate the watershed, we used all the pixels (or clusters or pixels larger than size 3) with RF confidence for membrane class $p_i^m < 0.01$. For superpixel clustering, we follow a context-aware agglomeration approach of~\cite{parag15b} that was designed to prevent under-segmentation by delaying some merge decisions during agglomeration. This agglomeration scheme first clusters the cytoplasm superpixels together using a superpixel boundary predictor and then absorbs the mitochondria bodies into the agglomerated cytoplasm regions based on their degree of inclusion. A superpixel boundary predictor for this setup considers the cell boundary as well as the border between mitochondria and cytoplasm as true boundaries and only the borders between over-segmented cytoplasm superpixels as false boundaries. 

Each boundary is represented by the statistical properties of the multiclass probabilities estimated by the pixel detector. The statistical properties include mean, standard deviation, 4 quartiles of the predictions generated for the data locations on the boundary, two regions it separates as well as the differences of these region statistics~\cite{parag15b}. All of these features can be updated in constant time after a merge -- a property which improves the efficiency of the segmentation algorithm substantially. The affinity values between two suprepixel boundaries were computed by the same formula  used for pixel classification. 
\begin{figure*}[t]
\vspace{-0.4cm}
\begin{center}
\subfigure[\scriptsize Split-VI test vol1]{\includegraphics[width=0.43\columnwidth, height=0.38\columnwidth]{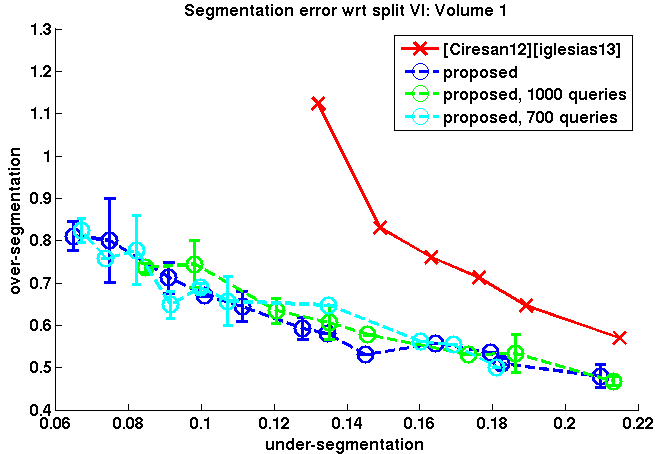}\label{F:VOL1VI}}
~~\subfigure[\scriptsize split-RE test vol1]{\includegraphics[width=0.43\columnwidth, height=0.38\columnwidth]{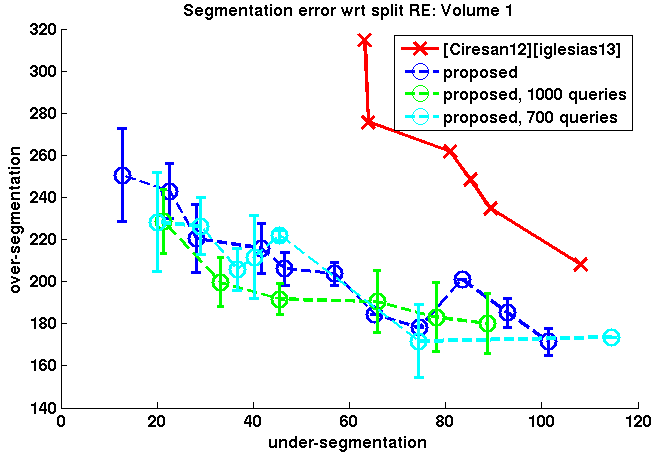}\label{F:VOL1RE}}
\qquad  \subfigure[\scriptsize Split-VI test vol2]{\includegraphics[width=0.43\columnwidth, height=0.38\columnwidth]{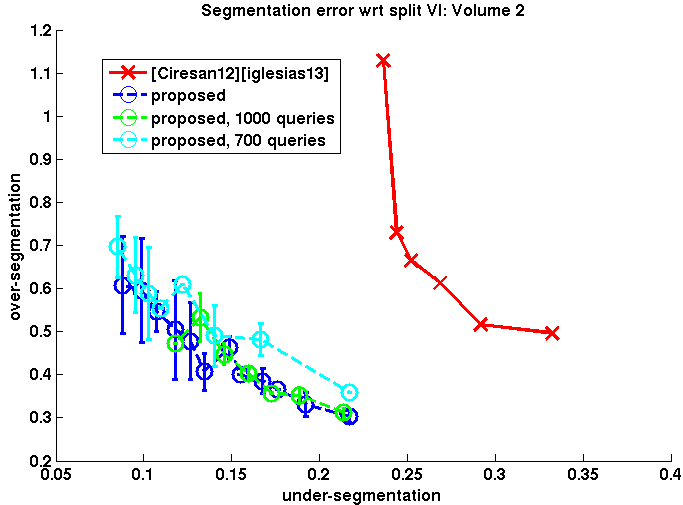}\label{F:VOL2VI}}
~~\subfigure[\scriptsize Split-RE test vol2]{\includegraphics[width=0.43\columnwidth, height=0.38\columnwidth]{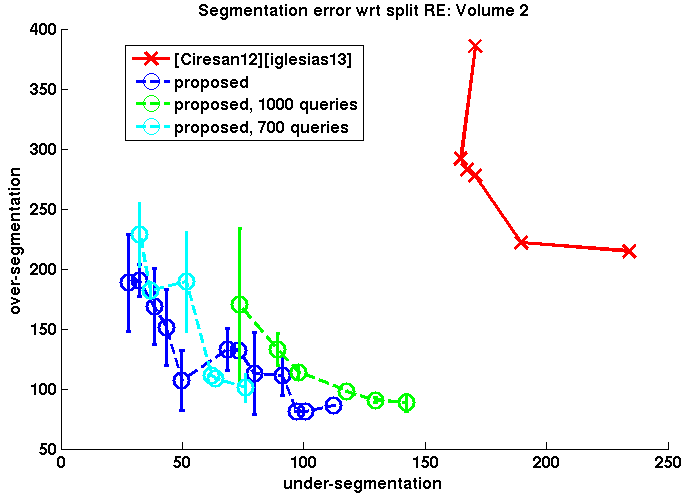}\label{F:VOL2RE}}
\vspace{-0.1cm}
\caption{\scriptsize Quantitative evaluation of competing methods on two FIBSEM test volumes. Left and right pairs of plots show the split-VI and split-RE errors of two methods on volume 1 and 2 respectively. }\label{F:FIBSEM_QUANT}
\end{center}
\vspace{-0.7cm}
\end{figure*}
\subsection{Result on 3D segmentation}\label{S:RESULT3D}
We have tested our algorithm for 3D volume segmentation on Focused Ion Beam Serial Electron Microscopy (FIBSEM) isotropic images collected from  fruit fly retina with a resolution of $10\times 10\times10$nm. One $250^3$ volume and two $520^3$ volumes were used as training and test datasets respectively.  The proposed algorithm does not need an exhaustive pixel-level groundtruth. However, for this particular experiment, instead of presenting queries to an annotator, we read off their labels from a noisy pixel level groundtruth generated earlier for another study~\cite{parag15b}. Each of the segmentation tasks, namely pixel classification, over-segmentation and subsequent context-aware agglomeration were performed in 3D. 

The performance of our algorithm was compared against a combination of~\cite{ciresan12} and~\cite{iglesias13} that has been one of the top scorer of the SNEMI 3D segmentation challenge 2013 (\url{http://brainiac2.mit.edu/SNEMI3D}). The neural net for pixel prediction was trained with the same techniques described in~\cite{ciresan12}\cite{giusti13}. In order to further improve the quality of the probability maps, the outputs on rotated images were averaged together~\cite{ciresan13}. The watersheds were generated in the same manner as those of the proposed method and then the agglomeration technique of~\cite{iglesias13} was applied for superpixel clustering.

We report the under- and over-segmentation errors separately because under-segmentation is costlier than the other in terms of manual correction. Given a groundtruth, $GT$, and a segmentation, $SG$, split versions of variance of information (VI)~\cite{meila03} and Rand Error (RE)~\cite{jain11} were selected for performance evaluation. For split-VI, the over and under-segmentation are quantified by the conditional entropy  $H(\text{GT} ~|~ \text{SG})$ and $H(\text{SG} ~|~ \text{GT})$ respectively. The over-segmentation  and under-segmentation quantities in Rand Error are the ratios of pixel pairs within same cluster in GT but different cluster in SG and vice versa. 

The proposed algorithm has been trained and applied 6 times to assess its consistency. In each training pass, we randomly subsampled a set of pixels from the whole training set so that the weight matrix $W$ used in label propagation contains $\sim 0.5\%$ nonzero values and still fit in the available memory. The remaining parameters of the proposed active learning scheme are fixed to initial set size $= 4000$ (1000 each class), query set size $=10$, number of queries $=800$ for all the experiments reported in this paper. For the superpixel boundary learning, the parameters are set for all experiments to initial set size $= 3.5\%$ of total number of boundaries, query set size $= 10$, number of total boundaries labeled $= 15\%$ of all examples ($10000 \sim 140000$ in total). With our current implementation, the computation of pixel and superpixel training scheme needed around 24 hours on a 32 and 16 core cluster node respectively.

In Figure~\ref{F:FIBSEM_QUANT}, we plot the split versions of error measures: x and y axes correspond to under- and over-segmentation errors respectively. Ideally, a segmentation algorithm should attain an error rate of 0, and therefore be plotted at the origin of the graph.  For both the proposed and that of~\cite{ciresan12}\cite{iglesias13}, the points on the plot were calculated by varying the stopping point of the agglomeration algorithm. The curve corresponding to the proposed method is an average of performances on 6 trials. On the two FIBSEM test volumes, the proposed algorithm (blue -o-, cyan and green curves are explained later) consistently produced lower false merge errors than that (red -x-) of~\cite{ciresan12}\cite{iglesias13} at the same over-segmentation error level. 

\begin{figure}
\vspace{-0.4cm}
\begin{center}
\subfigure[\scriptsize Plane 50]{\includegraphics[width=0.32\columnwidth, height=0.32\columnwidth]{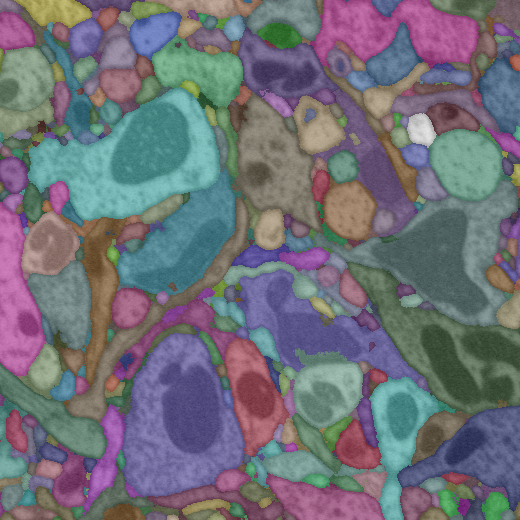}\label{F:PROP_QUAL50}}
\subfigure[\scriptsize Plane 220]{\includegraphics[width=0.32\columnwidth, height=0.32\columnwidth]{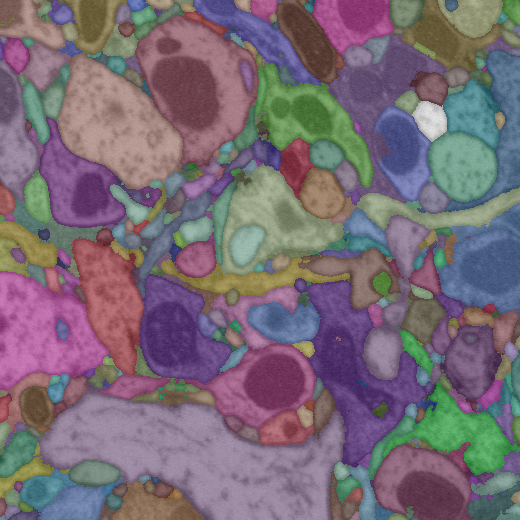}}
\subfigure[\scriptsize Plane 484]{\includegraphics[width=0.32\columnwidth, height=0.32\columnwidth]{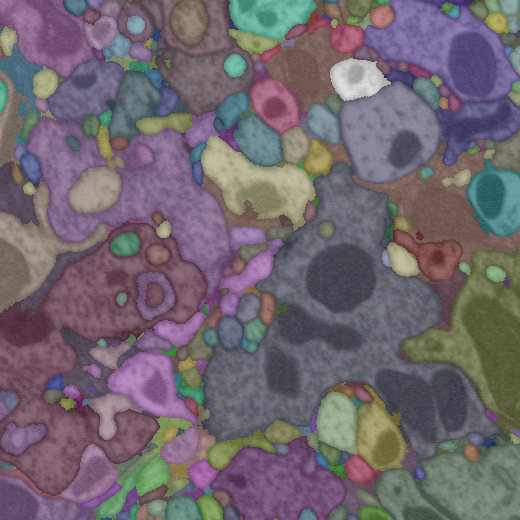}\label{F:PROP_QUAL484}}
\vspace{-0.1cm}
\caption{\scriptsize \subref{F:PROP_QUAL50}-\subref{F:PROP_QUAL484}:Result of the proposed method on  the same three slices as displayed in Figure~\ref{F:INTRO}. The output contains no false merges of significant size. }\label{F:PROPOSED_QUAL}
\end{center}
\vspace{-0.7cm}
\end{figure}

The combined methods of~\cite{ciresan12}\cite{iglesias13} generally attained high quality segmentation in most areas of the test volumes. However, because they do not emphasize on the membrane class for training, their outputs were vulnerable to false merges near relatively weaker membranes. In Figure~\ref{F:INTRO}, we have displayed the false merge generated by~\cite{ciresan12}\cite{iglesias13} operating at agglomeration threshold 0.15 (highest point on the red curve of Figure~\ref{F:FIBSEM_QUANT}\subref{F:VOL2VI}) on test volume 2. Segmentation produced by the proposed method did not reproduce this or any other false merges of similar size; the output of our method is shown in Figure~\ref{F:PROPOSED_QUAL}\subref{F:PROP_QUAL50}-\subref{F:PROP_QUAL484} for the same three planes. The qualitative results from the proposed method was generated with an agglomeration threshold of 0.3 (halfway in the blue curves of Figure~\ref{F:FIBSEM_QUANT}\subref{F:VOL2VI}). In both these images, the segmented regions are overlaid on the raw data with random color. Adjacent regions with same color may not always imply they are merged. The qualitative outputs on the two test volumes and a python script to visualize them can be found at \url{https://www.dropbox.com/sh/35x0z6md064yo88/AAAbH6JUwAwDKITDNnSsVEKga?dl=0}. A video of the output is also uploaded to youtube at \url{https://www.youtube.com/watch?v=osJtSJ8CSO4}. 

\begin{figure*}
\vspace{-0.4cm}
\begin{center}
\subfigure[\scriptsize Membr.from~\cite{ciresan12}]{\includegraphics[width=0.28\columnwidth, height=0.28\columnwidth]{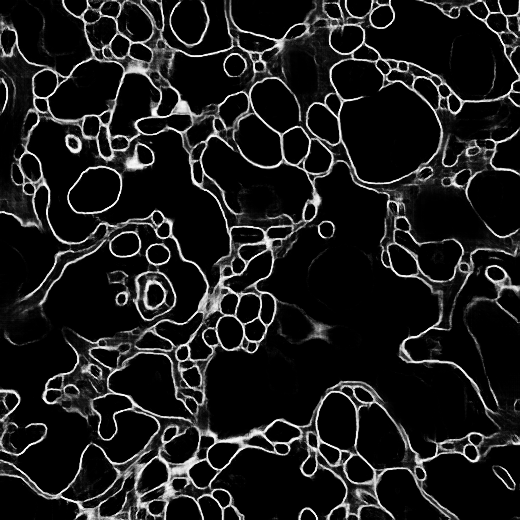}\label{F:DET_DAN}}
\subfigure[\scriptsize Prop. membr.]{\includegraphics[width=0.28\columnwidth, height=0.28\columnwidth]{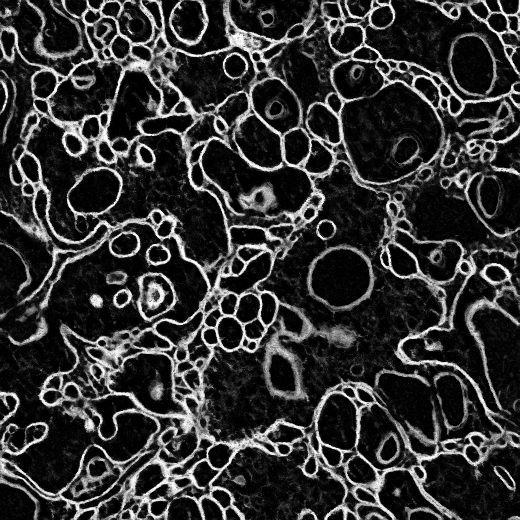}\label{F:DET_PROP}}
\subfigure[\scriptsize Prop. mitochond.]{\includegraphics[width=0.28\columnwidth, height=0.28\columnwidth]{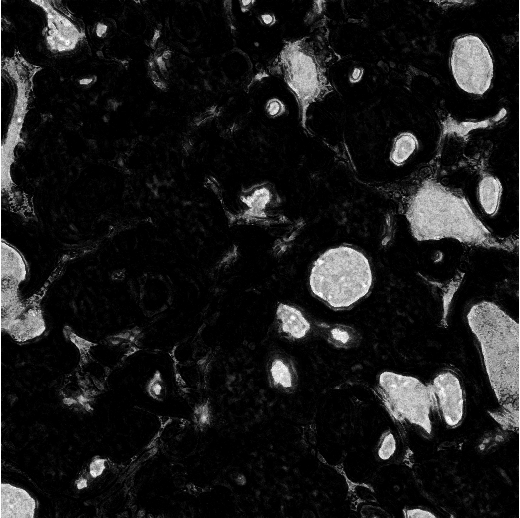}\label{F:DET_MITO}}
\subfigure[\scriptsize $\%$ pixel $p_i^m < 0.01$]{\includegraphics[width=0.39\columnwidth, height=0.33\columnwidth]{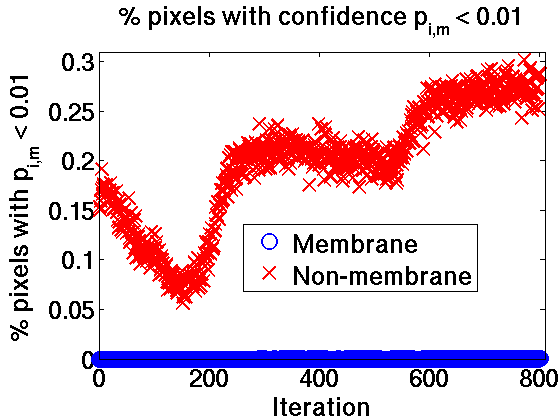}   \label{F:ZEROPROB}}
\subfigure[\scriptsize SP  accuracy]{\includegraphics[width=0.4\columnwidth, height=0.33\columnwidth]{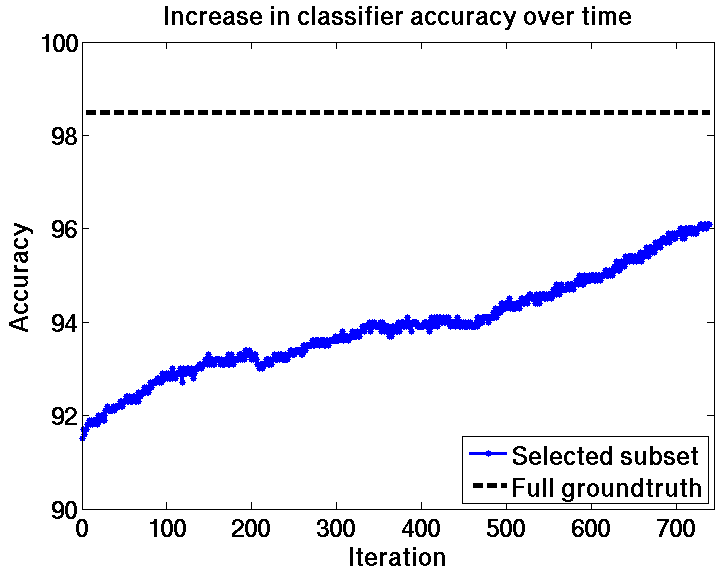}\label{F:SP_ACC}}
\subfigure[\scriptsize Error in SP query set]{\includegraphics[width=0.4\columnwidth, height=0.33\columnwidth]{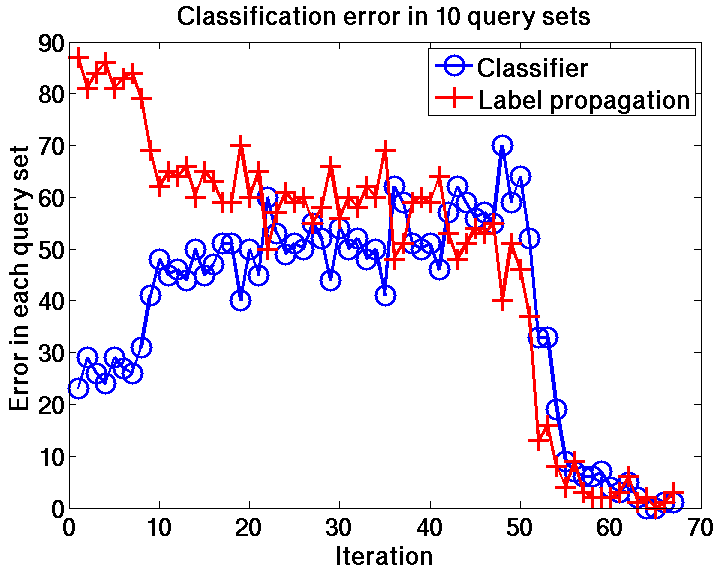}\label{F:SP_QUERY}}
\vspace{-0.1cm}
\caption{\scriptsize ~\subref{F:DET_DAN}-\subref{F:DET_MITO}: Pixel predictor confidences on Plane 484. ~\subref{F:DET_DAN} by~\cite{ciresan12},~\subref{F:DET_PROP} and~\subref{F:DET_MITO} by proposed method for boundary and mitochondria class respectively.~\subref{F:ZEROPROB}: percentage of pixels with $p_i^m < 0.01$ plotted against number of iterations.~\subref{F:SP_ACC}: increase in superpixel (SP) boundary classification accuracy with number of  iterations.~\subref{F:SP_QUERY}: prediction errors of the classifier and label propagation on every 10 query sets (100 samples) during superpixel boundary classification.}\label{F:FIBSEM_ANALYZE}
\end{center}
\vspace{-0.7cm}
\end{figure*}

In fact, both the test volumes were under-segmented in the watershed computed from~\cite{ciresan12}. The VI errors for under-segmentation for a watershed on~\cite{ciresan12} output were 0.132 and 0.236 respectively for two test volumes as opposed to 0.0188 and 0.0243 on average for those computed from our method. Such outcome may not be obvious from an examination of pixel probabilities computed by the proposed method and~\cite{ciresan12}; example predictions on Plane 484  are displayed in Figure~\ref{F:FIBSEM_ANALYZE}\subref{F:DET_DAN}-\subref{F:DET_MITO}. Indeed, the overall accuracy of our pixel detector is less than $90\%$ on samples whose labels are unknown to the active algorithm.  Although the deviation measure defined for active learning of pixel detection in Section~\ref{S:ACTIVE_PIXEL} enables the identification of misclassified locations, the gain in classification accuracy is not the prominent factor contributing to the low under-segmentation error of our technique.

The proposed pixel detection algorithm inherently minimizes the number of boundary pixels (and maximizes number of other types of pixels) receiving a confidence $p_i^m < 0.01$. Such an outcome is conducive to minimizing false merges in the consequent watershed method. In Figure~\ref{F:FIBSEM_ANALYZE}\subref{F:ZEROPROB} we plot the percentage of pixels of membrane (blue o) and other classes (red x) with $p_i^m< 0.01$ against the number of iterations. By construction, the algorithm starts with a very low, approximately $0.01\%$, of membrane pixels with $p_i^m < 0.01$. With the progression of the iterative updating of training examples, the proposed approach increases the percentage of other pixels with $p_i^m < 0.01$ while maintaining that for membrane pixels at the initial value. 

In case of the superpixel boundary classifier, however, the training scheme effectively reduces the classification error in distinguishing false boundaries from the correct ones. In Figure~\ref{F:FIBSEM_ANALYZE}\subref{F:SP_ACC}, we plot the increase in accuracy of the classifier being actively trained (blue curve) and that of the one learned from all examples (black dashed line) on test samples. The plot shows a steady performance improvement with query iterations (x-axis). Interestingly enough, the error rates of both the predictors, namely the label propagation and the classifier, on query sets of images drops to zero after a certain number of iterations as shown in Figure~\ref{F:FIBSEM_ANALYZE}\subref{F:SP_QUERY}. Such behavior has been observed in all the trials of superpixel boundary training and was utilized to determine a stopping criterion for training.

We have not reached to a point of zero error rates in query set for pixel classification. In order to test the sensitivity of the stopping criterion, we have plotted the error curves with 700, 800 and 1000 queries on Figure~\ref{F:FIBSEM_QUANT} in cyan, blue and green colors respectively.  The almost overlapping error curves suggest that the training converges in practice around 800 queries. After termination, the distribution of pixels in the whole dataset and those selected by the active semi-supervised algorithm are provided in Table~\ref{T:SELECTED}. Our algorithm selected more samples from the membrane class than one would choose by randomly sampling the same number of examples. 
\begin{table}
\scriptsize
\caption{\scriptsize Percentage of pixels in different classes in the whole dataset and the training set selected by the proposed pixel detection algorithm.}\label{T:SELECTED}\vspace{-0.3cm}
\begin{center}
\begin{tabular}{|c|c|c|c|c|}
\hline
		& cytoplasm & membrane & mitochon. & mito border \\
\hline
Whole dataset & 72.43 & 12.94 & 11.01 & 3.62 \\
Active selected & 52.09 & 30.59 & 14.78 & 2.54 \\
\hline
\end{tabular} 
\end{center}
\vspace{-0.4cm}
\end{table}

\begin{figure}
\vspace{-0.2cm}
\begin{center}
\subfigure{\includegraphics[width=0.32\columnwidth, height=0.3\columnwidth]{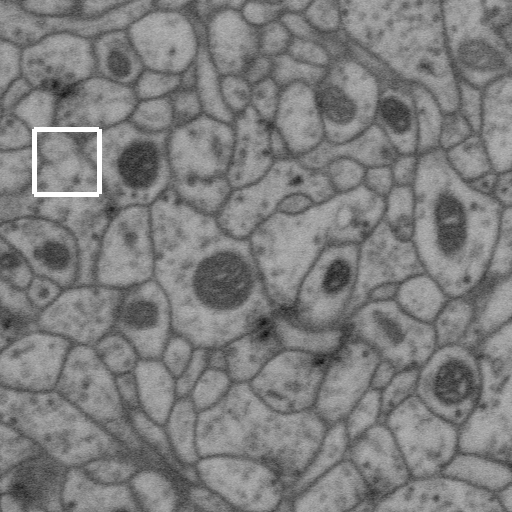}}
\subfigure{\includegraphics[width=0.32\columnwidth, height=0.3\columnwidth]{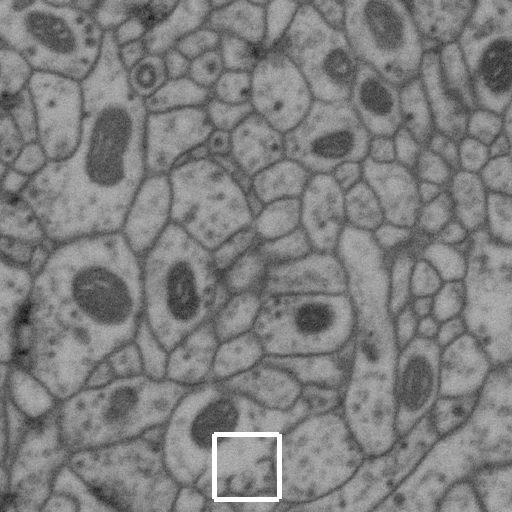}}
\subfigure{\includegraphics[width=0.32\columnwidth, height=0.3\columnwidth]{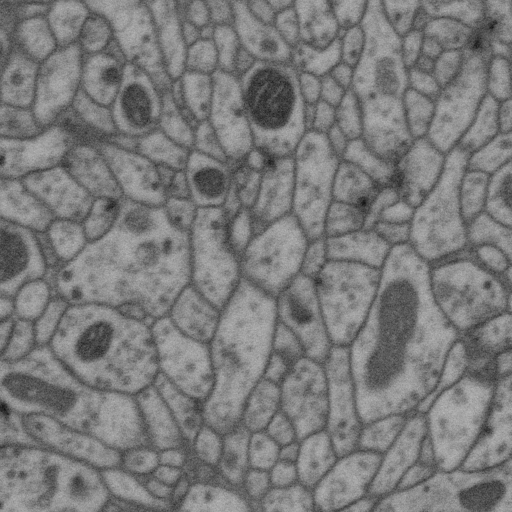}}
\vspace{-0.3cm}
\setcounter{subfigure}{0}
\subfigure[\scriptsize Plane 22]{\includegraphics[width=0.32\columnwidth, height=0.3\columnwidth]{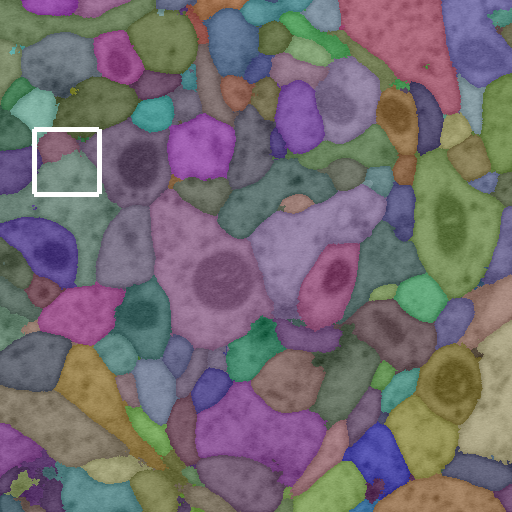}}
\subfigure[\scriptsize Plane 76]{\includegraphics[width=0.32\columnwidth, height=0.3\columnwidth]{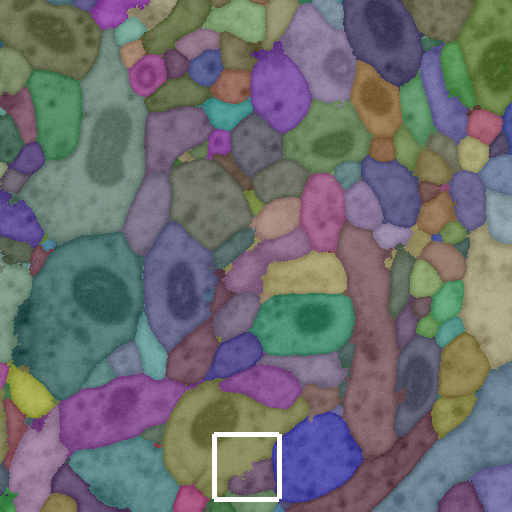}}
\subfigure[\scriptsize Plane 300]{\includegraphics[width=0.32\columnwidth, height=0.3\columnwidth]{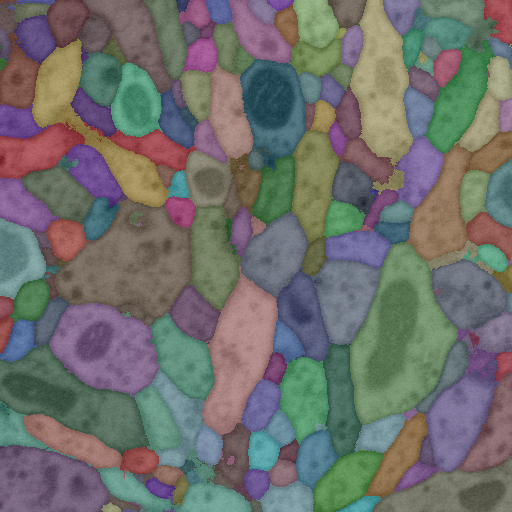}}
\vspace{-0.1cm}
\caption{\scriptsize Qualitative Result of the proposed method on  mushroom body FIBSEM data with exact same parameter. White boxes mark gaps in the membrane where the proposed method successfully avoided false merge.}\label{F:PROPOSED_QUAL_MB}
\end{center}
\vspace{-0.6cm}
\end{figure}

To further test the parameter sensitivity and robustness of our algorithm, we applied the proposed training with the exact same parameter on a $250^3$ FIBSEM volume from a different region (mushroom body) of fly brain and produced almost perfect segmentation on a separate $512^3$ mushroom body volume. Figure~\ref{F:PROPOSED_QUAL_MB} shows outputs on some of the planes,  note how the bias towards the membrane class of the proposed method resisted false merges on membrane gaps marked by white squares. Segmentation of all $512$ images can be found at \url{https://www.dropbox.com/sh/35x0z6md064yo88/AAAbH6JUwAwDKITDNnSsVEKga?dl=0}. The output is also uploaded to youtube at: \url{https://www.youtube.com/watch?v=mKnxxbQtN0g}.

\noindent \textbf{Performance of DAWMR~\cite{huang13dawmr} :} Our effort to test the capability of DAWMR~\cite{huang13dawmr}, which is an extended version of~\cite{jain10cvpr}, has not yet yielded results comparable to~\cite{ciresan12}\cite{iglesias13}. We attempt to analyze the reason behind such performance in the following text. 

The authors of~\cite{huang13dawmr} have kindly generated the affinity maps computed by the deep network for both our FIBSEM volumes. We computed the probability maps according to the authors' suggestion and applied~\cite{iglesias13} for superpixel clustering. At the same over-segmentation level as the proposed method (y-axis on Figure~\ref{F:FIBSEM_QUANT}), the result of DAWMR$+$~\cite{iglesias13} contained large incorrectly merged bodies in comparison to the  proposed method and the combination of~\cite{ciresan12}\cite{iglesias13}.

\begin{figure}[h]
\begin{center}
\subfigure{\includegraphics[width=0.47\columnwidth, height=0.47\columnwidth]{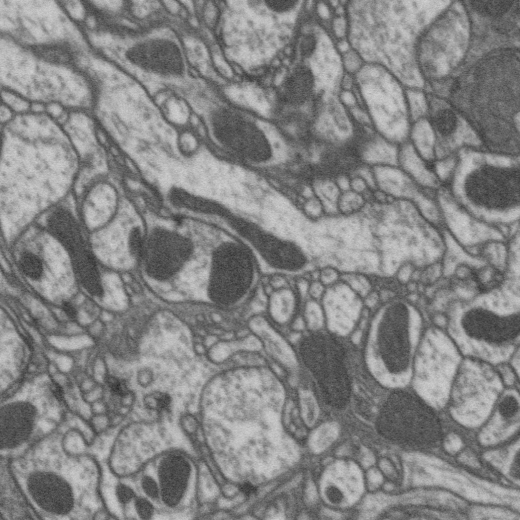}}~
\subfigure{\includegraphics[width=0.47\columnwidth, height=0.47\columnwidth]{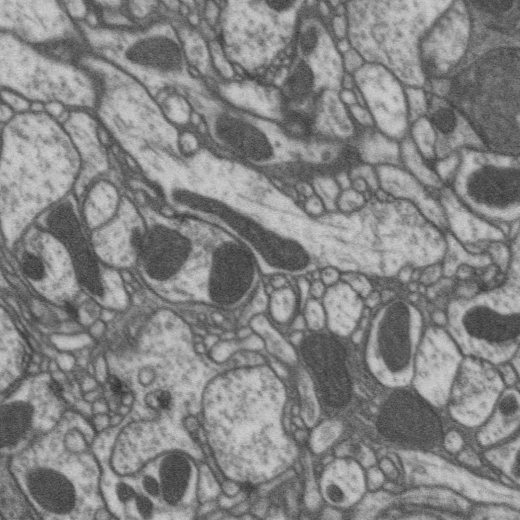}}
\setcounter{subfigure}{0}
\subfigure[\scriptsize Plane 339]{\includegraphics[width=0.49\columnwidth, height=0.5\columnwidth]{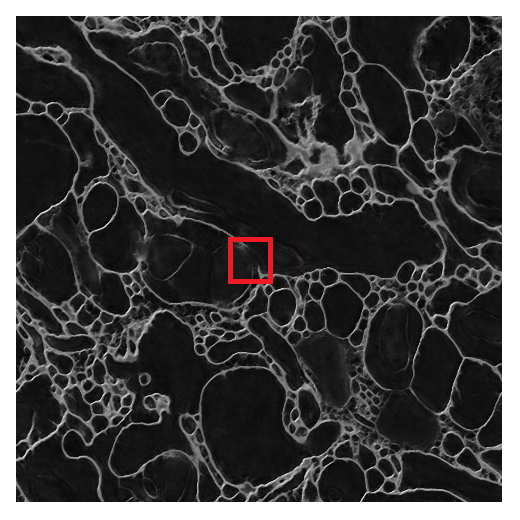}}
\subfigure[\scriptsize Plane 341]{\includegraphics[width=0.49\columnwidth, height=0.5\columnwidth]{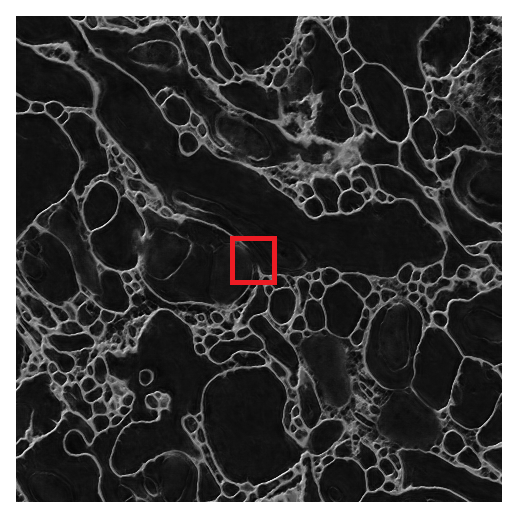}}
\end{center}\vspace{-0.3cm}
\caption{\scriptsize Top row: two planes on test volume. Bottom row: the membrane prediction of DAWMR~\cite{huang13dawmr} with weak confidences marked in red.}
\label{F:DAWMR_PRED}
\end{figure}

While investigating the reason behind this performance, we found that the pixel predictions of DAWMR for cell membrane fade away in several consecutive planes at multiple locations on the test volume. We show 3 such planes (339~341) in Figure~\ref{F:DAWMR_PRED}. These areas, some of which are marked in red on Figure~\ref{F:DAWMR_PRED}, are most probably responsible for joining two neurites inaccurately during the agglomeration. Recall that, the region statistics of pixelwise confidences are typically used as features for the superpixel boundary classifier~\cite{andres12}\cite{iglesias13}\cite{parag14}.  

\subsection{Result on 2D segmentation - ISBI 12}\label{S:RESULT2D}

We have also tested the proposed method for 2D segmentation on datasets provided for ISBI 2012 segmentation challenge (\url{http://brainiac2.mit.edu/isbi_challenge/home}). The challenge website provides a training set of 30 annotated images, generated by serial section Transmission Electron Microscopy (ssTEM) from the ventral nerve cord (VNC) of the Drosophila larva. We remind the user that an exhaustive groundtruth is not required by the proposed strategy because it automatically identifies the pixels and superpixel boundaries that are needed to be labeled by an annotator. For convenience of experimentation, and to incorporate some mistakes a human annotator would make in the active learning setting, we generated a noisy groundtruth by performing a watershed with all cell interior pixels marked as seeds and read off labels from this groundtruth.

A similar set of 30 images, without the groundtruth, was also provided for test purposes. The proposed method was applied on this dataset with the same number of samples and iteration for pixel classification as mentioned in Section~\ref{S:RESULT3D}. The number of examples utilized for superpixel boundaries is also similar to those stated in Section~\ref{S:RESULT3D}. In Table~\ref{T:RESULT_ISBI12}, we show the quantitative measures of performances of our method, that of~\cite{ciresan12} and another baseline algorithm that uses all pixels for training the pixel detector (Random Forest) and the technique of~\cite{iglesias13} for superpixel boundary training.  Since the groundtruth for the test dataset is not available, the split versions of VI and RE could not be computed. A qualitative inspection of the results (at \url{https://www.dropbox.com/sh/35x0z6md064yo88/AAAbH6JUwAwDKITDNnSsVEKga?dl=0}) suggests that the difference in error values between our method and those of~\cite{ciresan12} was most probably caused by over-segmentation. 

\begin{table}
\caption{\scriptsize Comparison of F-measure of Rand error provided by ISBI 2012 website.}\label{T:RESULT_ISBI12}\vspace{-0.2cm}
\begin{center}
\small
\begin{tabular}{|c|c|c|c|}
\hline
		& Proposed& ~\cite{ciresan12} & All+~\cite{iglesias13} \\
\hline
 error & 0.08 & 0.05 & 0.126 \\
\hline
\end{tabular} 
\end{center}
\vspace{-0.6cm}
\end{table}

For complete neuron reconstruction, the 2D segmentation results on anisotropic images -- such as those of ISBI 12 dataset -- need to be connected across planes by a linkage algorithm. The linkage algorithms have been shown to refine some false split errors, but cannot recover from false merges~\cite{funke12}\cite{vazquez11}. It is therefore rational (and may even be necessary) to prevent under-segmentation at a cost of small over-segmentation rate. This strategy will be more effective on difficult areas of EM volume characterized by broken or hazy membranes or dark cell regions. We downloaded 20 images from two different regions of the whole larva dataset (\url{http://fly.mpi-cbg.de/ })  and computed segmentation with predictors trained on the challenge data and the same set of parameters. For~\cite{ciresan12}, the output was generated by applying watershed after thresholding the pixel prediction values at $0.3$, same as that used to compute the winning entry of the ISBI 12 challenge.

\begin{figure}
\vspace{-0.2cm}
\begin{center}
\subfigure{\includegraphics[width=0.32\columnwidth, height=0.3\columnwidth]{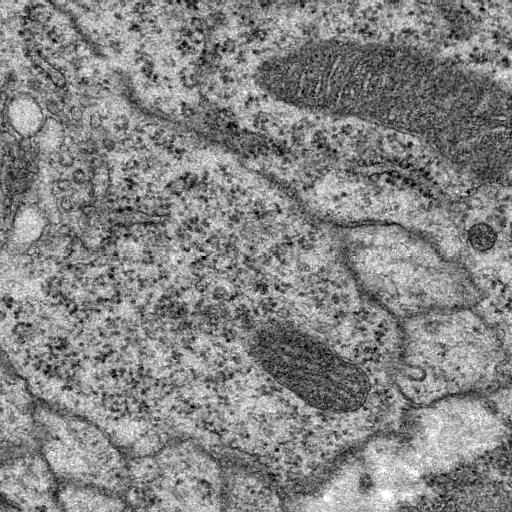}}
\subfigure{\includegraphics[width=0.32\columnwidth, height=0.3\columnwidth]{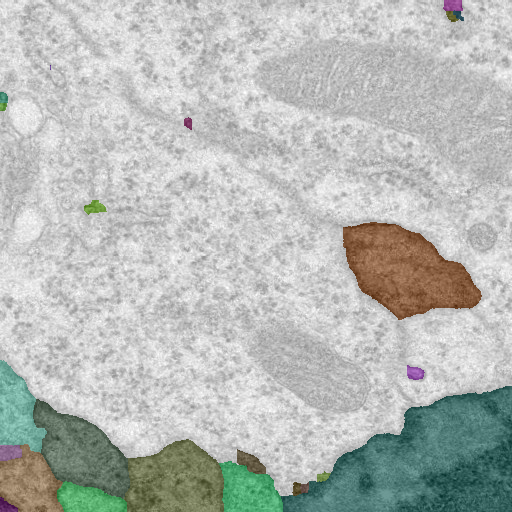}}
\subfigure{\includegraphics[width=0.32\columnwidth, height=0.3\columnwidth]{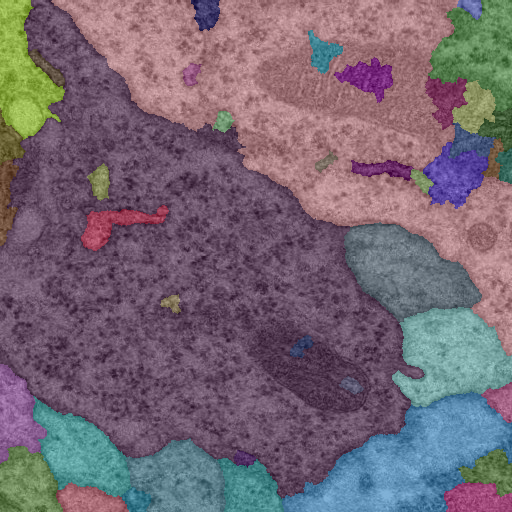}}
\vspace{-0.3cm}
\setcounter{subfigure}{0}
\subfigure[\scriptsize Input]{\includegraphics[width=0.32\columnwidth, height=0.3\columnwidth]{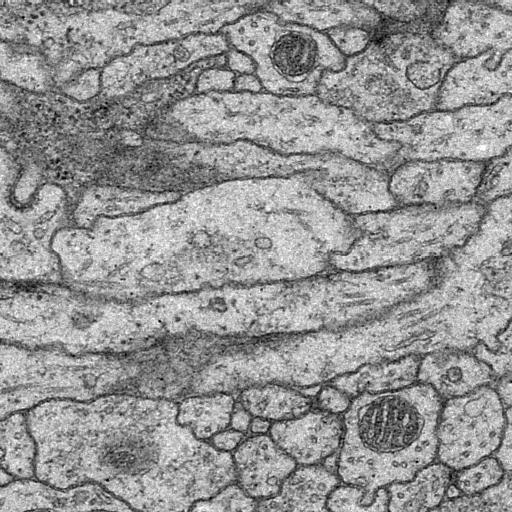}}
\subfigure[\scriptsize ~\cite{ciresan12}]{\includegraphics[width=0.32\columnwidth, height=0.3\columnwidth]{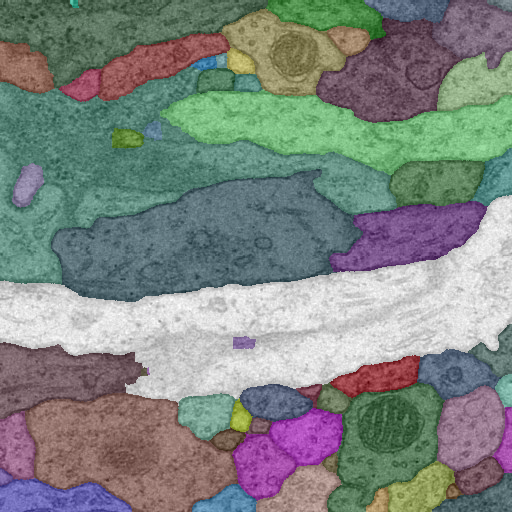}}
\subfigure[\scriptsize Proposed]{\includegraphics[width=0.32\columnwidth, height=0.3\columnwidth]{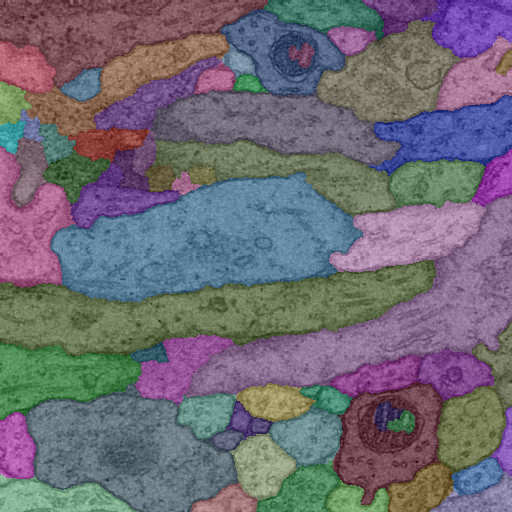}}
\vspace{-0.1cm}
\caption{\scriptsize Performance comparison with~\cite{ciresan12} on challenging regions of larva data. Regions highlighted in white (middle column) are falsely merged by watershed generated from~\cite{ciresan12}.}\label{F:LARVA2}
\vspace{-0.6cm}
\end{center}
\end{figure}

As Figure~\ref{F:LARVA2} demonstrates, the proposed method prevents most of the false merges  generated by~\cite{ciresan12} in these challenging areas and facilitates more accurate reconstruction through linkage algorithms like~\cite{funke12, vazquez11}.  An emphasis on learning the membrane class leads to a wall generally \lq higher\rq~ than those from~\cite{ciresan12} around watershed basins. Results on all the 20 images can be found at \url{https://www.dropbox.com/sh/35x0z6md064yo88/AAAbH6JUwAwDKITDNnSsVEKga?dl=0}.

\begin{figure}
\vspace{-0.3cm}
\begin{center}
\subfigure{\includegraphics[width=0.49\columnwidth, height=0.45\columnwidth]{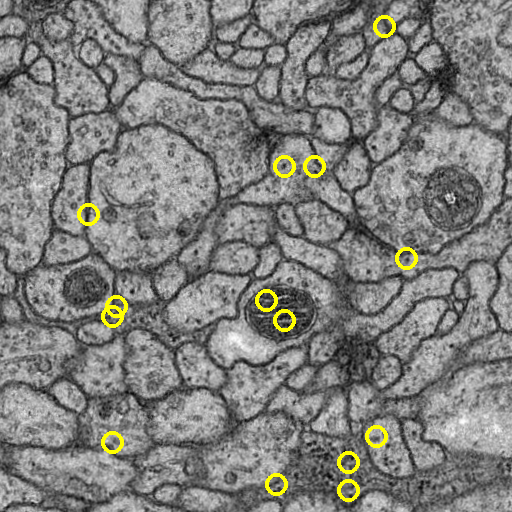}} 
\subfigure{\includegraphics[width=0.49\columnwidth, height=0.45\columnwidth]{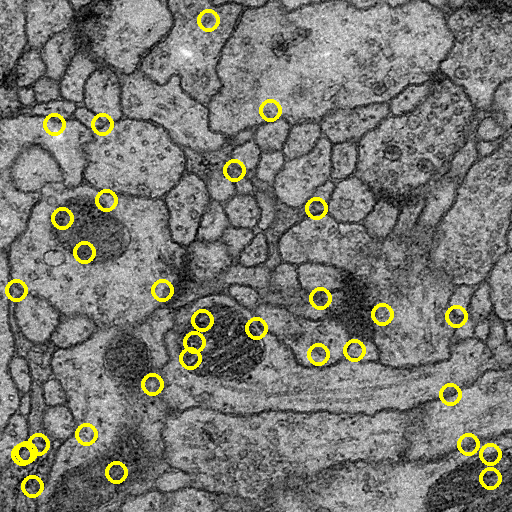}}
\vspace{-0.1cm}
\caption{\scriptsize Sample queries determined automatically by the proposed method. Note how the queries were placed at challenging locations on images such as patch between mitochondria and cell boundary, areas with darker shades.}
\label{F:ISBI_SELECTED}
\end{center}
\vspace{-0.7cm}
\end{figure}

Figure~\ref{F:ISBI_SELECTED} shows images with some pixel locations (circle centers) selected as queries by our active pixel training method. Recall that the query set consists of the challenging examples -- the locations where the estimation of the two techniques contradict each other. The regions covered by queries include patch between mitochondria and cell boundary, areas with darker shades. These regions often turn out to be misclassified (or receive low confidence) by a predictor trained in interactive setting of~\cite{ilastik11}. 

\begin{figure}
\vspace{-0.0cm}
\begin{center}
\subfigure{\includegraphics[width=\columnwidth, height=0.3\columnwidth]{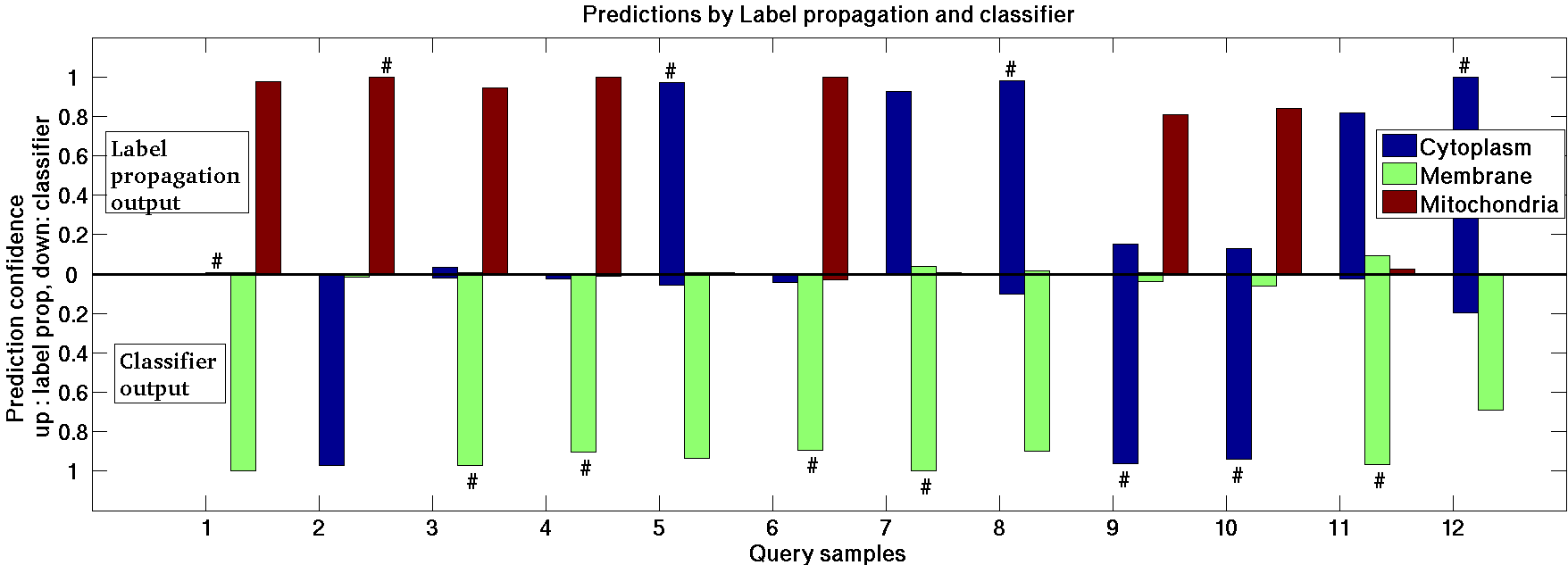}
\label{F:PRED_DEV}}
\vspace{-0.3cm}
\caption{\scriptsize The output confidences from the label propagation and the classifier on the first 12 examples returned as queries for 3 major classes: cytoplasm (blue), boundary (green) and mitochondria (brown). The top panel corresponds to label propagation predictions and the bottom shows that from the classifier (in the opposite direction, direction does not imply sign). The mark \# denotes the correct label of any particular sample.}
\label{F:ISBI_ANALYZE}
\end{center}
\vspace{-0.5cm}
\end{figure}

In Figure~\ref{F:ISBI_ANALYZE}, we show the output confidences from the label propagation algorithm and classifier on the first few samples selected as queries for three classes: cytoplasm (blue), membrane (green) and mitochondria (brown). The top and bottom panels correspond to the label propagation and RF respectively. The \# sign on top the bar shows the correct label for that particular sample. The plot shows how some samples misclassified by the RF classifier were correctly predicted by label propagation method and vice versa. Interestingly, the first sample was not detected accurately by any of the techniques. 

\section{Discussion}\label{S:DISCUSSION}
We have proposed a framework for training the necessary tools for an EM segmentation algorithm by acquiring some properties suitable for neural reconstruction. On one hand, the proposed method suggests a strategy to train without complete groundtruth by automatically selecting a small fraction of training examples. On the other hand, our algorithm is designed to minimize the false merge errors which are substantially more difficult to correct than the false split errors. The results demonstrate the merit of our method for neural reconstruction in comparison to the existing algorithms.


EM segmentation is a critical element of neural reconstruction process that led to high impact research in natural sciences, in particular neurobiology/neuroscience~\cite{takemura13}\cite{helmstaedter13a}. Our approach is designed to expedite multiple components of the overall reconstruction effort. For example, the neurobiologist who prepares the tissue sample currently relies only on visual inspection for sample quality assessment. A faster training method could assist the imaging expert to determine the optimal sample quality based on actual results rather than the raw images.

The authors of~\cite{takemura13} made an observation vital to the manual error correction step: screening $100\%$ of the segmentation result is impractical due to data size and is often redundant for extracting the underlying connectome. An intelligent strategy to automatically spot the areas needing correction, as proposed in~\cite{jones15proofread}, is perhaps essential for computing connectome from EM images. The presence of no or minimal under-segmentation is a prerequisite for applying methods such as ~\cite{jones15proofread}.       

With the increase of the size of brain region that the researchers ponder on reconstructing, it is anticipated that appearances (and therefore the feature distributions) of different regions of brain would vary considerably from one another.  An efficient approach for preparing the automated algorithms may be inevitable for scenarios where one must train different predictors for different regions of large datasets.

Finally, although the algorithm is modeled and tested primarily for EM reconstruction,  it has a potential to be applied in other domains. Techniques that use superpixels aggregation to produce the final segmentation, e.g.,~~\cite{amat14} for cell tracking in light microscopy, ~\cite{wang14histo} for blood cell segmentation,  can utilize our method for efficiency and performance improvement. 

{\small
\section*{Acknowledgment}
Toufiq Parag gratefully acknowledges Pat Rivlin, Chris Ordish, Corey Fisher for assistance in data preparation; Stuart Berg, Steve Plaza for software support; Gary Huang for computing DAWMR output; Stephan Saalfeld for providing the access to the complete larva dataset.
}

{\small
\bibliographystyle{ieee}
\bibliography{limited_gt_final}

\begin{thebibliography}{10}\itemsep=-1pt

\bibitem{amat14}
F.~Amat, W.~Lemon, D.~P. Mossing, K.~McDole, Y.~Wan, K.~Branson, E.~W. Myers,
  and P.~J. Keller.
\newblock {Fast, accurate reconstruction of cell lineages from large-scale
  fluorescence microscopy data}.
\newblock {\em Nat Meth}, 11(9):951--958, 2014.

\bibitem{amat13}
F.~Amat, E.~W. Myers, and P.~J. Keller.
\newblock Fast and robust optical flow for time-lapse microscopy using
  super-voxels.
\newblock {\em Bioinformatics}, 29(3):373--380, 2013.

\bibitem{andres12}
B.~Andres, T.~Kroeger, K.~Briggman, W.~Denk, N.~Korogod, G.~Knott, U.~Koethe,
  and F.~Hamprecht.
\newblock Globally optimal closed-surface segmentation for connectomics.
\newblock In {\em ECCV}. 2012.

\bibitem{meyer93}
S.~Beucher and F.~Meyer.
\newblock The {M}orphological {A}pproach to {S}egmentation : {T}he {W}atershed
  {T}ransformation.
\newblock {\em Mathematical Morphology in Image Processing}, pages 433--481,
  1993.

\bibitem{breiman01}
L.~Breiman.
\newblock Random forests.
\newblock {\em Machine Learning}, 45(1):5--32, Oct. 2001.

\bibitem{semi-super06}
O.~Chapelle, B.~Sch{\"o}lkopf, and A.~Zien, editors.
\newblock {\em Semi-Supervised Learning}.
\newblock MIT Press, 2006.

\bibitem{chklovskii10}
D.~B. Chklovskii, S.~Vitaladevuni, and L.~K. Scheffer.
\newblock Semi-automated reconstruction of neural circuits using electron
  microscopy.
\newblock {\em Current Opinion in Neurobiology}, 20(5):667--675, Oct. 2010.

\bibitem{chung96}
F.~R.~K. Chung.
\newblock {\em {Spectral Graph Theory (CBMS Regional Conference Series in
  Mathematics, No. 92)}}.
\newblock American Mathematical Society, Dec. 1996.

\bibitem{ciresan12}
D.~C. Ciresan, A.~Giusti, L.~M. Gambardella, and J.~Schmidhuber.
\newblock Deep neural networks segment neuronal membranes in electron
  microscopy images.
\newblock In {\em NIPS}, 2012.

\bibitem{ciresan13}
D.~C. Ciresan, A.~Giusti, L.~M. Gambardella, and J.~Schmidhuber.
\newblock Mitosis detection in breast cancer histology images with deep neural
  networks.
\newblock In {\em MICCAI}, volume~2, pages 411--418, 2013.

\bibitem{funke12}
J.~Funke, B.~Andres, F.~Hamprecht, A.~Cardona, and M.~Cook.
\newblock Efficient automatic 3d-reconstruction of branching neurons from em
  data.
\newblock In {\em CVPR}, 2012.

\bibitem{giusti13}
A.~Giusti, D.~C. Ciresan, J.~Masci, L.~M. Gambardella, and J.~Schmidhuber.
\newblock Fast image scanning with deep max-pooling convolutional neural
  networks.
\newblock In {\em ICIP}, 2013.

\bibitem{review09histo}
M.~Gurcan, L.~Boucheron, A.~Can, A.~Madabhushi, N.~Rajpoot, and B.~Yener.
\newblock Histopathological image analysis: A review.
\newblock {\em Biomedical Engineering, IEEE Reviews in}, 2:147--171, 2009.

\bibitem{helmstaedter13b}
M.~Helmstaedter.
\newblock Cellular-resolution connectomics: challenges of dense neural circuit
  reconstruction.
\newblock {\em Nature Methods}, 10(6):501--7, 2013.

\bibitem{helmstaedter13a}
M.~Helmstaedter, K.~L. Briggman, S.~C. Turaga, V.~Jain, H.~S. Seung, and
  W.~Denk.
\newblock {Connectomic reconstruction of the inner plexiform layer in the mouse
  retina}.
\newblock {\em Nature}, 500(7461):168--174, Aug. 2013.

\bibitem{huang13dawmr}
G.~B. Huang and V.~Jain.
\newblock Deep and wide multiscale recursive networks for robust image
  labeling.
\newblock {\em CoRR}, abs/1310.0354, 2013.

\bibitem{jain10cvpr}
V.~Jain, B.~Bollmann, M.~Richardson, D.~Berger, M.~Helmstaedter, Briggman,
  et~al.
\newblock Boundary learning by optimization with topological constraints.
\newblock In {\em CVPR}, 2010.

\bibitem{jain10opinion}
V.~Jain, S.~Seung, and S.~Turaga.
\newblock Machine that learn to segment images: a crucial technology for
  connectomics.
\newblock {\em Current opinion in Neurobiology}, 20:653--666, 2010.

\bibitem{jain11}
V.~Jain, S.~C. Turaga, K.~Briggman, M.~N. Helmstaedter, W.~Denk, and H.~S.
  Seung.
\newblock Learning to agglomerate superpixel hierarchies.
\newblock In {\em NIPS 24}, pages 648--656. 2011.

\bibitem{jones15proofread}
C.~Jones, T.~Liu, N.~W. Cohan, M.~Ellisman, and T.~Tasdizen.
\newblock Efficient semi-automatic 3d segmentation for neuron tracing in
  electron microscopy images.
\newblock {\em Journal of Neuroscience Methods}, 246(0):13 -- 21, 2015.

\bibitem{jones13sparse}
C.~Jones, T.~Liu, M.~Ellisman, and T.~Tasdizen.
\newblock Semi-automatic neuron segmentation in electron microscopy images via
  sparse labeling.
\newblock In {\em {ISBI}}, pages 1304--1307, 2013.

\bibitem{jurrus10mia}
E.~Jurrus, A.~R.~C. Paiva, S.~Watanabe, J.~R. Anderson, B.~W. Jones, R.~T.
  Whitaker, E.~M. Jorgensen, R.~Marc, and T.~Tasdizen.
\newblock Detection of neuron membranes in electron microscopy images using a
  serial neural network architecture.
\newblock {\em Medical Image Analysis}, 14(6):770--783, 2010.

\bibitem{kelley95}
C.~T. Kelley.
\newblock {\em Iterative Methods for Linear and Nonlinear Equations}.
\newblock Number~16 in Frontiers in Applied Mathematics. SIAM, 1995.

\bibitem{liu14}
T.~Liu, C.~Jones, M.~Seyedhosseini, and T.~Tasdizen.
\newblock A modular hierarchical approach to 3d electron microscopy image
  segmentation.
\newblock {\em Journal of Neuroscience Methods}, 226(0):88 -- 102, 2014.

\bibitem{liu12watershed}
T.~Liu, E.~Jurrus, M.~Seyedhosseini, M.~H. Ellisman, and T.~Tasdizen.
\newblock Watershed merge tree classification for electron microscopy image
  segmentation.
\newblock In {\em ICPR}. IEEE, 2012.

\bibitem{meila03}
M.~Meila.
\newblock Comparing clusterings by the variation of information.
\newblock In {\em COLT'03}, pages 173--187, 2003.

\bibitem{meila01}
M.~Meila and J.~Shi.
\newblock A random walks view of spectral segmentation.
\newblock In {\em {AISTATS} 2001}, 2001.

\bibitem{iglesias13}
J.~Nunez-Iglesias, R.~Kennedy, T.~Parag, J.~Shi, and D.~B. Chklovskii.
\newblock Machine learning of hierarchical clustering to segment 2d and 3d
  images.
\newblock {\em PLoS ONE}, 8(8), 08 2013.

\bibitem{parag15b}
T.~Parag, A.~Chakraborty, S.~Plaza, and L.~Scheffer.
\newblock A context-aware delayed agglomeration framework for electron
  microscopy segmentation.
\newblock {\em PLoS ONE}, 10(5), 2015.

\bibitem{parag14}
T.~Parag, S.~Plaza, and L.~Schefher.
\newblock Small sample learning of superpixel classifier for em segmentation.
\newblock In {\em MICCAI}, 2014.

\bibitem{ilastik11}
C.~Sommer, C.~Straehle, U.~Koethe, and F.~A. Hamprecht.
\newblock "ilastik: Interactive learning and segmentation toolkit".
\newblock In {\em ISBI}, 2011.

\bibitem{takemura13}
S.-Y. Takemura et~al.
\newblock A visual motion detection circuit suggested by {Drosophila}
  connectomics.
\newblock {\em Nature}, 500(7461):175--181, 2013.

\bibitem{vazquez11}
A.~Vazquez-Reina, M.~Gelbart, D.~Huang, J.~Lichtman, E.~Miller, and H.~Pfister.
\newblock Segmentation fusion for connectomics.
\newblock In {\em ICCV}, 2011.

\bibitem{wang14histo}
J.~Wang, J.~D. MacKenzie, R.~Ramachandran, and D.~Z. Chen.
\newblock Identifying neutrophils in h{\&}e staining histology tissue images.
\newblock In {\em {MICCAI}}, 2014.

\bibitem{zhu02tech}
X.~Zhu and Z.~Ghahramani.
\newblock {Learning from labeled and unlabeled data with label propagation},
  2002.
\newblock Tech Report CMU-CALD-01-107, School of Computer Science, Carnegie
  Mellon University.

\bibitem{zhu03}
X.~Zhu, J.~Lafferty, and Z.~Ghahramani.
\newblock {Combining Active Learning and Semi-Supervised Learning Using
  Gaussian Fields and Harmonic Functions}.
\newblock In {\em {ICML 2003 workshop on The Continuum from Labeled to
  Unlabeled Data in Machine Learning and Data Mining}}, 2003.

\end{thebibliography}
}

\end{document}